\documentclass[10pt,twocolumn,letterpaper]{article}

\usepackage[pagenumbers]{cvpr} % To force page numbers, e.g. for an arXiv version

\usepackage{graphicx}
\usepackage{amsmath}
\usepackage{amssymb}
\usepackage{booktabs}

\usepackage[pagebackref,breaklinks,colorlinks]{hyperref}

\usepackage[capitalize]{cleveref}
\crefname{section}{Sec.}{Secs.}
\Crefname{section}{Section}{Sections}
\Crefname{table}{Table}{Tables}
\crefname{table}{Tab.}{Tabs.}

\def\cvprPaperID{*****} % *** Enter the CVPR Paper ID here
\def\confName{CVPR}
\def\confYear{2022}

\begin{document}

\title{Broken News: Making Newspapers Accessible to Print-Impaired}

\author{Vishal Agarwal \hspace{10pt} Tanuja Ganu \hspace{10pt} Saikat Guha\\
Microsoft Research, India \\
% Bengaluru, India \\
{\tt\small \{vishal.agarwal, tanuja.ganu, saikat\}@microsoft.com}
% \and
}
\maketitle

\begin{abstract}
    Accessing daily news content still remains a big challenge for people with print-impairment including blind and low-vision due to opacity of printed content and hindrance from online sources. In this paper, we present our approach for digitization of print newspaper into an accessible file format such as HTML. We use an ensemble of instance segmentation and detection framework for newspaper layout analysis and then OCR to recognize text elements such as headline and article text. Additionally, we propose EdgeMask loss function for Mask-RCNN \cite{maskrcnn} framework to improve segmentation mask boundary and hence accuracy of downstream OCR task. Empirically, we show that our proposed loss function reduces the Word Error Rate (WER) of news article text by 32.5 \%.
\end{abstract}

\section{Introduction}
\label{sec:intro}

% * Lack of accessible quality news for BLV. Digitize Newspapers.
% * Segmenting articles then OCR has unacceptably high WER. Due to noisy ground-truth.
% * Fixing GT still has poor WER. Due to edges.
% * Root cause is IoU loss not being sensitive to edges. Nor other suitable loss functions.
% * Contributions:
% *   i. Propose new loss function, metric. 
% *  ii. Method to clean up noisy ground-truth.
% * iii. Trained model with 75% reduction in error relative to default.

Access to daily news content is challenging for blind, low-vision, and otherwise print-disabled individuals~\cite{manik}. Online news websites are not screen-reader friendly being cluttered with menus, ads, popups, and sidebars, reading which consumes minutes before getting to the actual article content, and having to repeat that for each and every article. News on social media is often low-quality. And print newspapers are not accessible.

We attempted to solve the accessible news problem by applying computer vision to print newspapers, specifically segmenting articles and performing OCR, however, we discovered existing approaches resulted in poor quality. The primary problem was the lack of ground-truth at scale suitable for training. We used automated means to collect additional ground-truth, which was lower quality than expected. Section~\ref{sec:experiments:data} details challenges and the data cleaning process we ultimately devised for a high-quality newspaper article segmentation dataset.

Training a segmentation model on this dataset and performing OCR still resulted in high Word and Character Error Rates (WER/CER). We diagnosed this high error rate to two contributors. First was the inherent WER/CER of the OCR engine, which was fairly low due to the nature of newspaper print. The predominant factor was small errors in segmentation boundary, sometimes of even a single pixel, which would render characters on the \emph{edges} of the article unrecognizable. This is fundamentally due to the cross-entropy mask loss function, which forms the basis of all segmentation models, being extremely non-sensitive to the \emph{boundary} pixels as long as there is high area overlap. Section~\ref{sec:methodology:edgemask} details the problem with the mask loss in this regard, and our enhanced loss function -- EdgeMask -- that addresses this problem.

Overall, this work makes three contributions. First, we present our approach to scale ground-truth for newspaper digitization. Second, we propose the EdgeMask loss function for news article segmentation to help improve accuracy of downstream OCR tasks. And finally, we report on experimental results of a 32.5\% reduction in WER in newspaper digitization.

\section{Background}
\label{sec:background}

% * efforts to make pdf accessible to BLV
% * Google tried to digitize newspaper to index and make it searchable on its search engine
% * independent work on extracting visual content and newspaper article segementation
% * OCR system for historical documents
% * IoU-based loss function modification for specific applications

Some of the past work has been done for extracting visual content and digitizing historical newspaper for searching and archival purposes\cite{newspinflib, googlenewspaper, navigator}. The Google Newspaper Search project \cite{googlenewspaper} was one of the large-scale efforts to digitize newspapers, index it and make the article contents discoverable via search engine. \cite{navigator} proposed a pipeline for extracting and searching visual content from historic print newspaper scans. A fully convolutional segmentation approach has been applied by \cite{fcnarticle} and \cite{dnnnews} to extract content blocks from newspaper images. 

Besides newspaper, there had been similar efforts on digitizing general document PDF and making it accessible. \cite{historyocr} developed an OCR system for historical documents. \cite{pdfaccessible} proposed various techniques to improve accessibility of scanned PDFs to visually impaired.

\begin{figure*}[h]
  \centering
%   \fbox{\rule{0pt}{2in} \rule{0.9\linewidth}{0pt}}
   \includegraphics[width=0.8\linewidth]{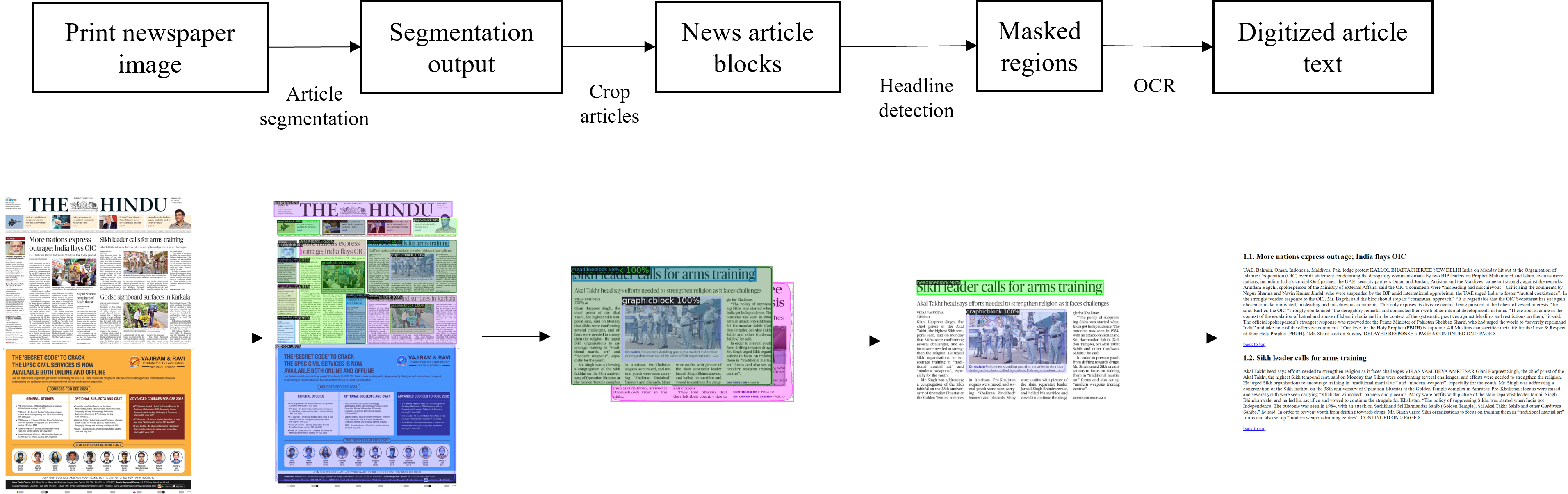}
   \caption{Illustration of the newspaper digitization pipeline}
   \label{fig:newspipeline}
\end{figure*}

In the newspaper space, there has been number of effort to digitize entire or some elements of newspaper for search related purposes. In the general print document space, there has been work on digitizing as well as making it accessible to people with vision impaired. To our knowledge, there has been no exact work on making print newspaper accessible to print-impaired people including blind and low-vision. 

\section{Methodology}
\label{sec:methodology}

% * Newspaper digitization pipeline description
% * Modifying loss function
% * WER/CER is a better metric than mAP

This section details the newspaper digitization process and the problem with mask loss of Mask-RCNN-based architecture. It also describes our proposed EdgeMask loss function for mitigating digitization issues and the metric used for evaluation.

\subsection{Newspaper digitization}
\label{sec:methodology:newsdigi}

The process of digitizing a print newspaper is divided into 3 steps. First, understanding the layout of the newspaper for segmenting news article blocks. Secondly, detecting elements like headline, images, graphical illustration and paragraph text within a news article. And third, applying OCR to extract text and structuring the text based on the semantics of the content. In a newspaper page, there are different block of contents like news articles, ads, illustrations, header and footer. Among these, the most important is news article content and we digitize only such content at present.

The news articles are cropped out using an instance segmentation model, based on the Mask RCNN architecture. A typical article contains a headline, article text and images or graphical illustrations. After segmentation, the headlines and graphical illustration are detected and masked as white pixels. The remainder is the article text which is then extracted using an OCR engine. Similarly, the headline is also extracted using OCR by cropping the headline-detected region. The digitized article is generated in a markdown format with headline as a header and article text as paragraph text. A digitized accessible newspaper is generated by collating all articles in a single file with all headlines in a table of content on the top of the page. A sample output is shown in Fig. \ref{fig:htmlnews}. Fig. \ref{fig:newspipeline} illustrates the end-to-end pipeline for digitization. Section \ref{sec:experiments:training} details about the complete process of training both news article segmentation model and the headline detection model.

\subsection{EdgeMask loss}
\label{sec:methodology:edgemask}

We used the Mask-RCNN framework and adopted the architecture for training the news article segmentation model. We observed high word error rate and character error rate after OCR. This was primarily due to text getting chopped off at the segmentation boundaries. The original loss function of Mask-RCNN is defined as a multi-task loss as shown in \ref{maskrcnnloss}
\begin{equation}
  L = L_{cls} + L_{box} + L_{mask}
  \label{maskrcnnloss}
\end{equation}

\noindent where $L_{cls}$ is the classification loss, $L_{box}$ is the bounding box regression loss and $L_{mask}$ is the average binary cross-entropy loss for all the pixels in the mask. This formulation works well for general object instance segmentation because even if the predictions are slightly off by few pixels at the boundary, the end result of identifying an object is not affected. On the other hand, this is very sensitive for newspaper digitization. If the segmentation boundary is smaller than the ground-truth boundary at the edges, then it would miss out on the bottom lines of text and if the boundary is slightly extended, then it would include text from neighboring articles. Both of these scenarios are undesirable for our end task of digitizing the news article contents.

We tackled this problem by proposing a new loss function EdgeMask loss for the Mask RCNN framework. The mask loss function $L_{mask}$ is modified to improve accuracy of the segmentation towards the boundary. We put more weights for pixels at the boundary of the mask than the pixels at the center and hence penalizing the boundary pixels more for misclassification. The EdgeMask loss is defined as follows
\begin{equation}
\label{eq:edgemask}
    L_{EdgeMask} = \frac{1}{m^2} [\sum_{\substack{(i,j) \in I}}^{} h_{ij} + \lambda \sum_{(p,q) \in B}^{} h_{pq}]
\end{equation}

% \begin{equation}
% \label{eq:edgemask}
%     L_{EdgeMask} = \frac{1}{m^2} [\sum_{(i,j) \in interior points}^{m-k}\mathop{}_{\mkern-5mu ij} h_{ij} + \lambda \sum_{(p,q) \in boundary points}^{m}\mathop{}_{\mkern-5mu pq} h_{pq}]
% \end{equation}

where $B$ is the set of points at boundary of mask with window of size $k$, $I$ is the set of points interior to the mask, $\lambda$ is the weights for the pixels at boundary and $h_{ij}$ is the binary cross-entropy for a pixel $(i,j)$. For each pixel in the RoI of size $m$x$m$, the binary cross-entropy is defined as shown in equation \ref{eq:pixelbce}.
\begin{equation}
\label{eq:pixelbce}
    h_{ij} = - [y_{ij} \log \hat{y_{ij}} + (1-y_{ij}) \log (1-\hat{y_{ij}})]
\end{equation}

\subsection{Metrics}
\label{sec:methodology:metrics}

All standard metrics for object detection or segmentation task depend on the fundamental concept of average precision (AP). It is computed by calculating the area under the precision-recall curve where TP or FP is calculated by considering some IoU threshold between the predicted boxes and the ground-truth boxes. Some of the common metric include AP, AP@.75, AP@[.5:.05:.95] for each classes and similarly mAP averaged over all classes.

While the IoU-based metrics work fine for generic object detection or segmentation tasks, this metric was not found to be suitable for our application. Our objective is to segment each article region independently. As mentioned in section \ref{sec:methodology:edgemask}, the digitization problem is sensitive to predictions at the boundary of the mask and typical IoU metrics doesn't capture the same. Considering a scenario when the predicted mask is just a few pixels shorter than the ground-truth mask which results in missing out on the last 1-2 lines of the article text. Although the IoU is still very high, the end digitized text would be meaningless because of the incomplete text. This calls for an alternative metric that could be representative for the task at hand.

In our experiments, we use Character Error Rate (CER) and Word Error Rate (WER) for measuring the correctness of the end digitized text. Using the predicted segmentation mask, we apply OCR on it and compare the predicted text with the ground truth text. 

\section{Experiments}
\label{sec:experiments}

% * Data challenges - Lack of ground-truth and data cleaning process
% * Training Mask-RCNN with modified loss and multiple hyperparameters
% * Final result for 2 segmentation models and overall digitization result

This section talks about the dataset, the process to generate good quality ground-truth and the training process of the vision models for our digitization pipeline. We also discuss results with various hyperparameters and illustrate segmentation output result as well as the end output of digitized accessible news.

\subsection{Data}
\label{sec:experiments:data}
The data consists of 5600 images of print newspapers and ground-truth of two types - 1) segmentation masks for 5 categories of elements - article, ad, header, headline and graphical illustration, and 2) string of text for each article in the newspaper. The primary challenge related to the data was lack of readily-available ground-truth. So we scraped the internet to find newspaper images with any kind of bounding box or segmentation annotations.

For the majority of data, either the ground-truth segmentation didn't exist or they were of low quality -- overlapping masks and not spanning the actual area of content. Besides, there were lack of label for each mask corresponding to different category of elements. We improved the ground-truth segmentations and added labels by combination of manual labeling, heuristics and labeling by vision models.

First, we create a high-quality dataset by manually labeling 100 images which is then used to fine-tune a Faster-RCNN object detection model, pre-trained on the COCO dataset. Using the fine-tuned model, we generated labels and bounding box annotations for all the newspaper images. The quality of the annotation was further improved by using text bounding box outputs from an OCR region. We take an intersection of the predicted bounding boxes from fine tuned Faster-RCNN model and the text bounding boxes from an OCR engine to get a tighter, high-quality rectangle fit for all news articles. We use a heuristic based on font size and bounding box results from OCR region to distinguish between the labels -- headlines and article text. With a combination of such heuristics, very small-scale manual labeling and large-scale vision-based model labeling, we generated good quality ground-truth for the digitization task.

\begin{figure}
  \centering
  \begin{subfigure}{0.4\linewidth}
    % \fbox{\rule{0pt}{2in} \rule{.9\linewidth}{0pt}}
    \includegraphics[width=1.1\linewidth]{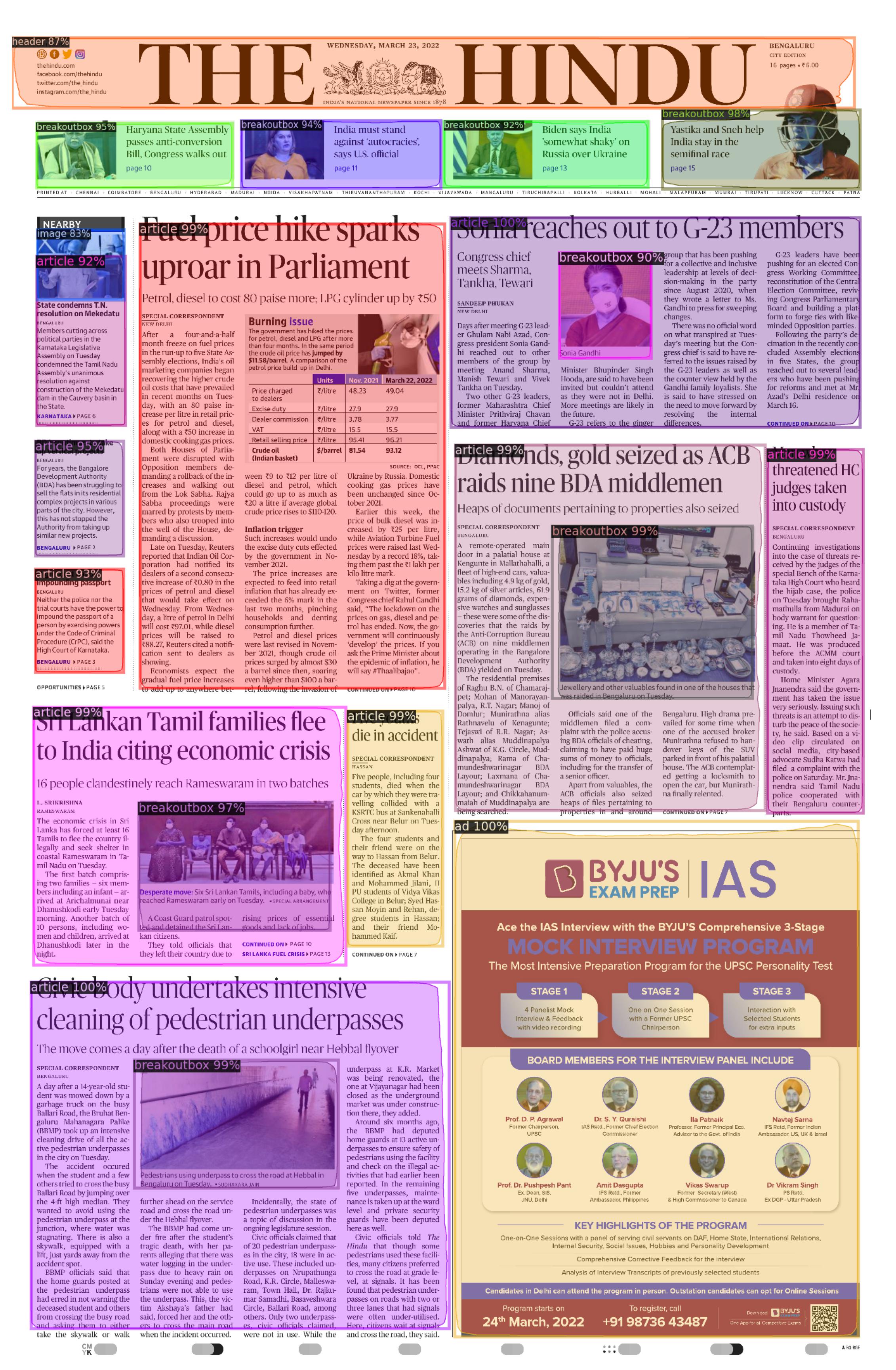}
    \caption{vanilla Mask-RCNN loss}
    \label{fig:short-a}
  \end{subfigure}
  \hspace{0.3in}
%   \hfill
  \begin{subfigure}{0.4\linewidth}
    % \fbox{\rule{0pt}{2in} \rule{.9\linewidth}{0pt}}
    \includegraphics[width=1.1\linewidth]{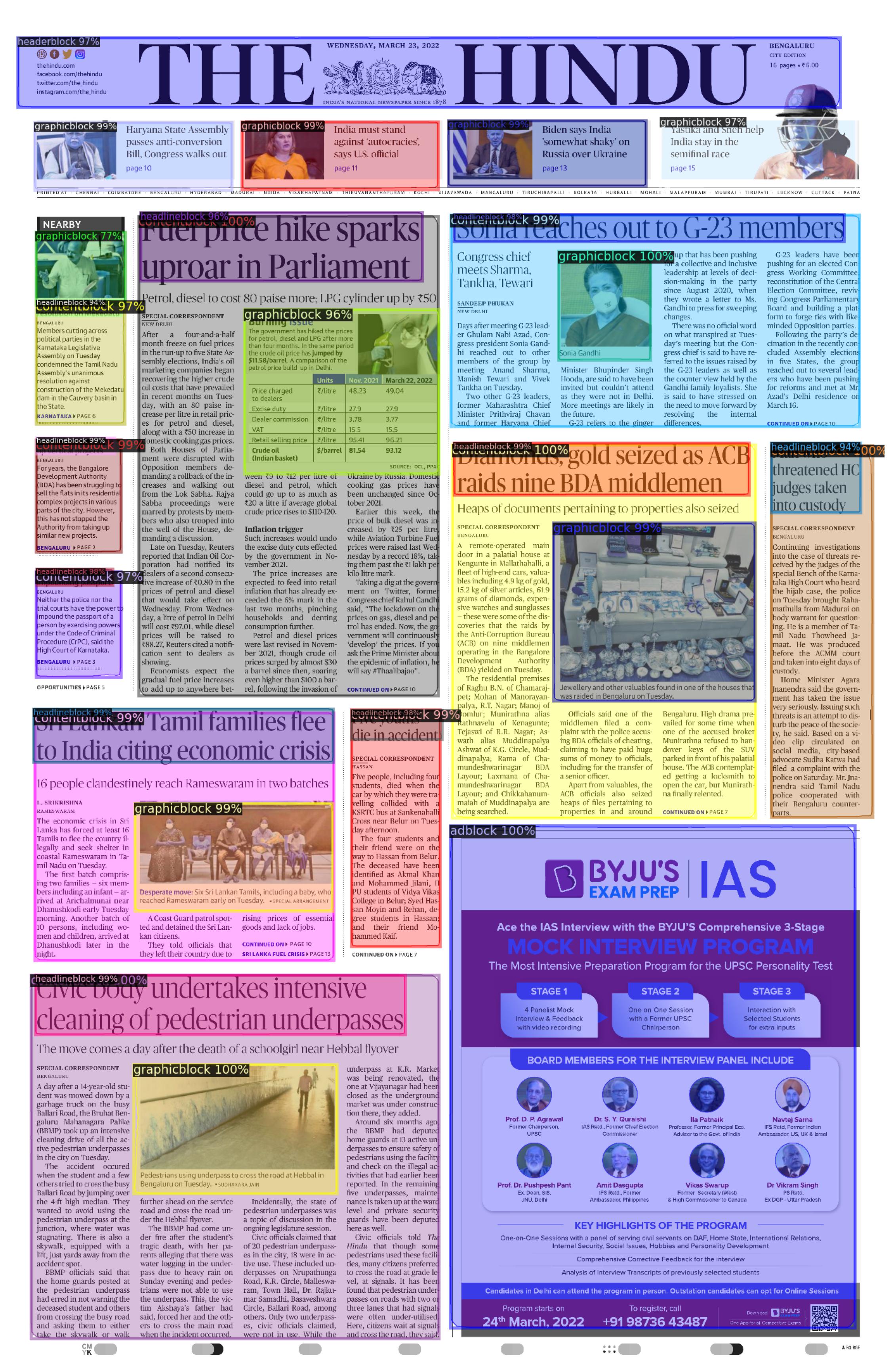}
    \caption{EdgeMask loss, $\lambda=100$}
    \label{fig:short-b}
  \end{subfigure}
  \caption{This figure illustrates the segmentation output with the original mask loss and improved boundaries with EdgeMask loss}
  \label{fig:short}
\end{figure}

\subsection{Training}
\label{sec:experiments:training}
The details of the digitization pipeline is as follows - given a newspaper page, an instance segmentation model crops out all newspaper articles. For each cropped article, we pass it through a detection model to detect 2 classes of elements - headlines and images. These elements are then masked and the article image is then feed through OCR engine to extract article text. Similarly, the headline region is cropped and digitized using OCR. The news article is hence reconstructed from digitized headline and article text. 

We trained two vision models for our digitization pipeline -- 1) an instance segmentation model based on Mask-RCNN architecture for segmenting news articles, 2) an object detection model based on Faster-RCNN architecture to detect headlines and images within news article blocks.

We adopted the Mask-RCNN framework from the Detectron2 \cite{detectron2} model zoo for identification and segmentation of news article blocks from images of print newspaper. The architecture was based on ResNet50 feature pyramid network (FPN) base config and the loss function criterion was modified with our proposed EdgeMask loss. The model was trained for 50k iterations on 5600 print newspaper images.  We experimented with multiple values of hyperparameter $\lambda$=1, 30, 100, 1000 and 10000 for the EdgeMask loss and found $\lambda=100$ to be the optimal. 

After cropping the news article, we use a Faster-RCNN object detection model to detect headlines and images. This model was adopted from the Detectron2 model zoo as well and trained with 260,000 news article block images.

We evaluated the performance of our pipeline using the metric CER and WER. Table \ref{tab:cerwer} shows WER and CER values for different values of $\lambda$ when computed for boundary text. Table \ref{tab:apscore} shows various AP-based metrics for both vision models.

\begin{table}
  \centering
  \begin{tabular}{c|c|c}
    \toprule
    $\lambda$ & WER \%   & CER \% \\
    \midrule
    1 & 8.89 & 3.19 \\
    10 & 8.10 & 2.82 \\
    30 & 7.17 & 2.35 \\
    \bf{100} & \bf{6.00} & \bf{1.77} \\
    1000 & 15.66 & 4.81 \\
    \bottomrule
  \end{tabular}
  \caption{Word and Character Error Rate (WER/CER) of boundary text for various $\lambda$}
  \label{tab:cerwer}
\end{table}

% \begin{table}
%   \centering
%   \begin{tabular}{c|cc|cc}
%     \toprule
%     $\lambda$ & \multicolumn{2}{c}{WER\%}   & \multicolumn{2}{c}{CER\%} \\
%     \midrule
%     & full & boundary & full & boundary \\ 
%     1 & 3.86 & 8.89 & 1.15 & 3.19 \\
%     10 & 3.28 & 8.10 & 1.09 & 2.82 \\
%     30 & 3.18 & 7.17 & 1.01 & 2.35 \\
%     \bf{100} & \bf{2.13} & \bf{6.00} & \bf{0.73} & \bf{1.77} \\
%     1000 & 7.72 & 15.66 & 2.51 & 4.81 \\
%     \bottomrule
%   \end{tabular}
%   \caption{Word and Character Error Rate (WER/CER) of full article text and text sampled at boundaries for various $\lambda$}
%   \label{tab:cerwer}
% \end{table}

\begin{table}
  \centering
  \setlength{\tabcolsep}{4pt}
  \begin{tabular}{@{}c|cccccc@{}}
    \toprule
    \footnotesize Model & \footnotesize AP & \footnotesize AP@$.50$ & \footnotesize AP@$.75$ & \footnotesize AP$_{m}$ & \footnotesize AP$_l$ \\
    \midrule
    \footnotesize Article segmentation & \footnotesize 86.11 & \footnotesize 93.46 & \footnotesize 91.85 & \footnotesize 30.30 & \footnotesize 86.71 \\ 
    \footnotesize Headline detection & \footnotesize 84.31 & \footnotesize 91.79 & \footnotesize 89.53 & \footnotesize 62.63 & \footnotesize 85.53 \\
    \bottomrule
  \end{tabular}
  \caption{AP score}
  \label{tab:apscore}
\end{table}

\begin{figure}
  \centering
  \begin{subfigure}{0.4\linewidth}
    % \fbox{\rule{0pt}{2in} \rule{.9\linewidth}{0pt}}
    \includegraphics[width=1.1\linewidth]{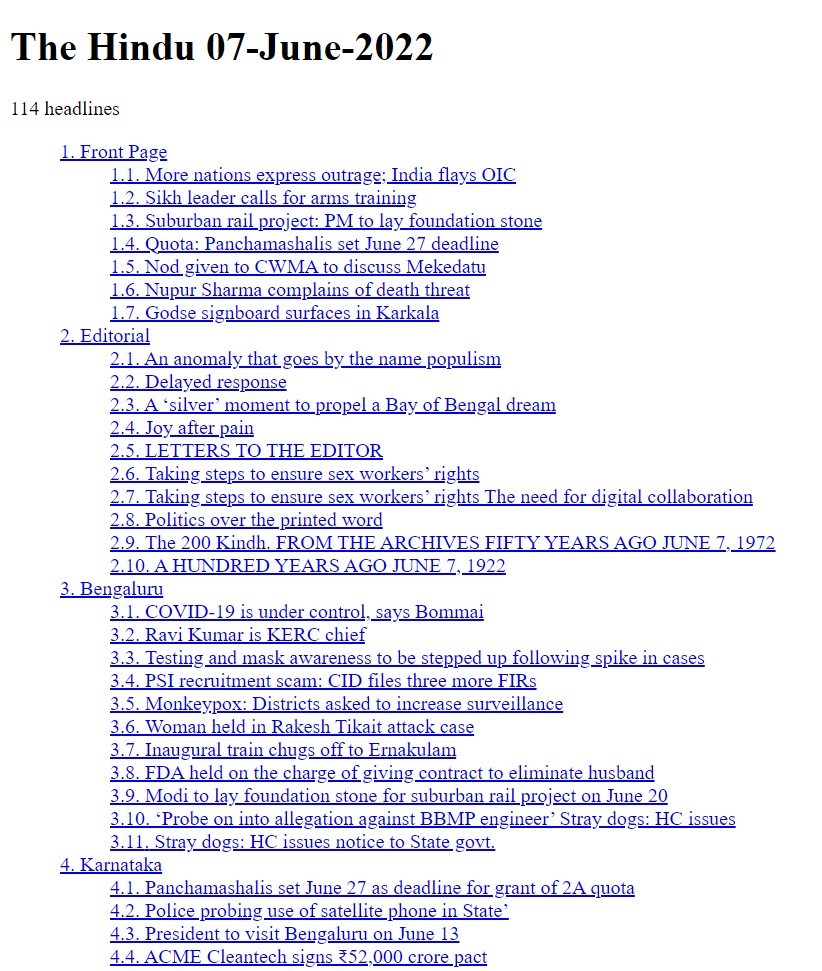}

    % \caption{vanilla Mask-RCNN loss}
    \label{fig:short-a}
  \end{subfigure}
  \hspace{0.3in}
%   \hfill
  \begin{subfigure}{0.4\linewidth}
    % \fbox{\rule{0pt}{2in} \rule{.9\linewidth}{0pt}}
    \includegraphics[width=1.1\linewidth]{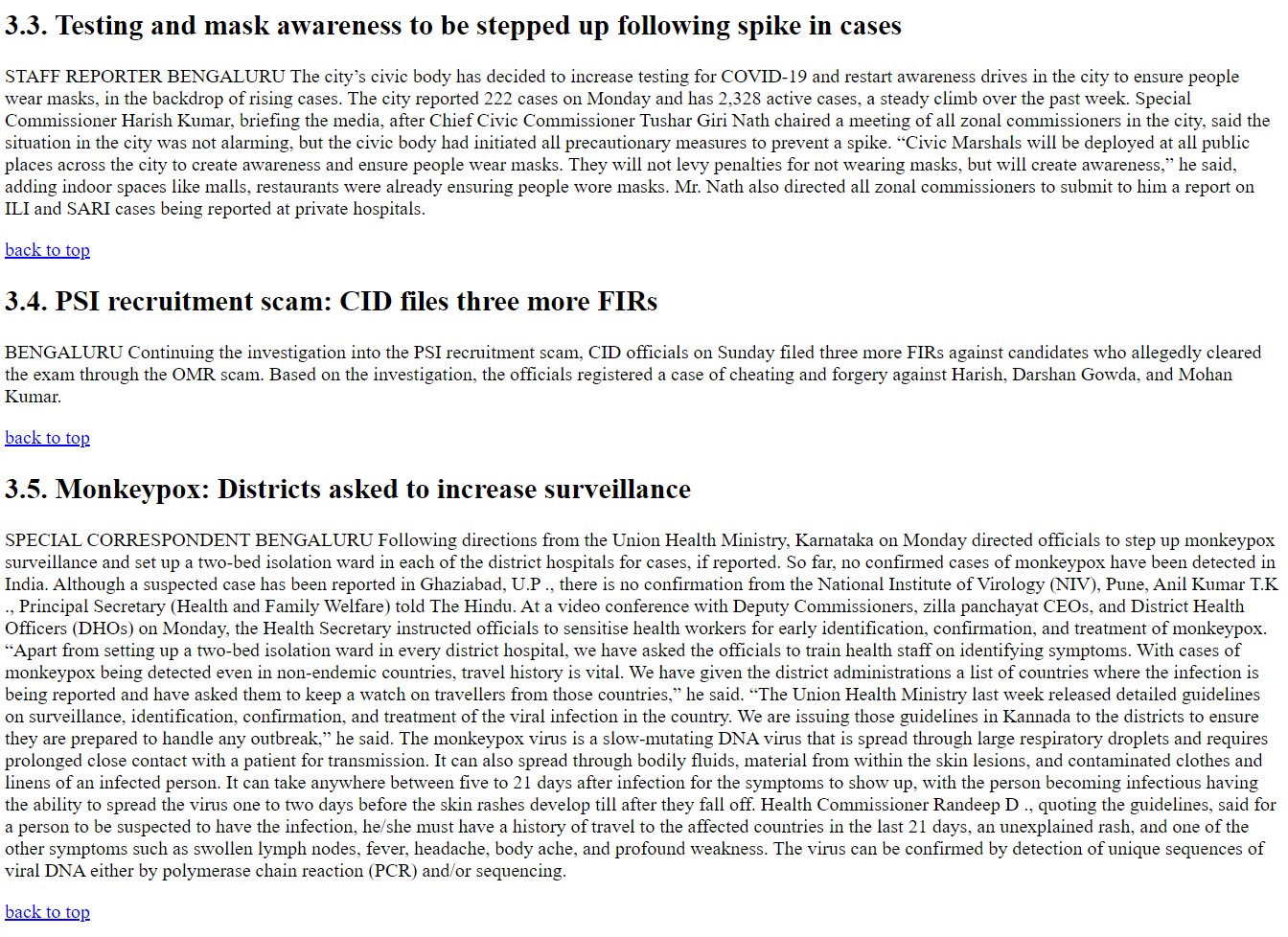}    
    % \caption{EdgeMask loss}
    \label{fig:short-b}
  \end{subfigure}
  \caption{Digitized newspaper in HTML}
  \label{fig:htmlnews}
\end{figure}

\section{Discussion}
\label{sec:discussion}

% * connect articles split across two pages
% * auto alt-text for images, generate descriptions for various plots, illustrations, comics
% * newspaper is just a start. Work towards making all types of printed content more accessible

In Fig. \ref{fig:htmlnews}, we show the HTML file generated from our newspaper digitization pipeline. The top of the page contains information about the newspaper such as date and name of the newspaper, followed by a table of contents. The digitized version was generated by collating text from all the article blocks independently. In few newspapers, we noticed that some articles are split across two pages, providing only a gist of information on the first page. Such cases will be considered as two different articles in our pipeline and in future, we plan to tackle this as such a design is fairly common across newspapers.

There are various kinds of graphical elements or visual illustrations found in a newspaper, e.g. images, tables, plots, comics, etc. Although, in our pipeline, we detect any such illustration but ignore for digitization. In order to make a newspaper truly digitally accessible, one of the most important direction to work towards is to automatically generate alt-text for any visual illustration and meaningful descriptions for plots, tables and comics \cite{alt1, alt2, alt3, alt4}.

\section{Conclusion}
\label{sec:conclusion}

In this work, we focus towards digitizing and hence making print newspaper accessible to print-disabled individuals. We adopted state-of-the-art computer vision techniques and proposed modifications to aid to our use-case. We proposed EdgeMask loss function for Mask-RCNN framework to improve segmentation mask prediction at the boundary. This significantly improved the end digitized news article texts with a reduction of 32.5\% word error rate compared to the vanilla loss criterion. We also shared how we tackled data challenges such as lack of ground-truth using a combination of manual labelling, heuristics and labeling by a vision model. This is our first stepping stone towards our broader goal to make all kinds of printed content accessible.

%%%%%%%%%%%
\iffalse

\fi

%%%%%%%%% REFERENCES
{\footnotesize
\bibliographystyle{ieee_fullname}
\bibliography{egbib}
}

\end{document}